%% file: main.tex
\documentclass{article}

\usepackage[nonatbib, final]{neurips_2019}

\usepackage[utf8]{inputenc} 
\usepackage[T1]{fontenc}    
\usepackage{hyperref}       
\usepackage{url}            
\usepackage{booktabs}       
\usepackage{amsfonts}       
\usepackage{nicefrac}       
\usepackage{microtype}      
\usepackage{graphicx}
\usepackage{subfigure} 
\usepackage{diagbox}

\usepackage{amsmath}
\usepackage{dsfont}
\usepackage{caption}

\title{
An Adaptive View of Adversarial Robustness from Test-time Smoothing Defense
}

\author{
  Chao Tang\thanks{Equal contribution}  ,
  Yifei Fan\footnotemark[1]  , \textnormal{and}
  Anthony Yezzi\\
  School of Electrical and Computer Engineering\\
  Georgia Institute of Technology\\
  Atlanta, GA 30308 \\
  \texttt{\{njtangchao96, yifei\}@gatech.edu} \\
}

\begin{document}

\maketitle

\begin{abstract}
The safety and robustness of learning-based decision-making systems are under threats from adversarial examples, as imperceptible perturbations can mislead neural networks to completely different outputs. 
In this paper, we present an adaptive view of the issue via evaluating various test-time smoothing defense against white-box untargeted adversarial examples. 
Through controlled experiments with pretrained ResNet-152 on ImageNet, we first illustrate the non-monotonic relation between adversarial attacks and smoothing defenses.
Then at the dataset level, we observe large variance among samples and show that it is easy to inflate accuracy (even to 100\%) or build large-scale (i.e., with size ~$10^4$) subsets on which a designated method outperforms others by a large margin.
Finally at the sample level, as different adversarial examples require different degrees of defense, the potential advantages of iterative methods are also discussed.    
We hope this paper reveal useful behaviors of test-time defenses, which could help improve the evaluation process for adversarial robustness in the future.%
\end{abstract}

\section{Introduction}
\input{sections/intro.tex}

\section{Related work} \label{sec:literature}
\input{sections/literature.tex}
\section{Methodology} \label{sec:methodology}
\input{sections/methodology.tex}
\section{Experiments} \label{sec:experiment}
\input{sections/experiment.tex}
\section{Discussion} \label{sec:discussion}
\input{sections/discussion.tex}

\subsection*{Acknowledgement}
This work was funded in part by Army Research Office grant number ARO W911NF-18-1-0281 and Google Cloud Platform Research Credit award.
\newpage
{\small
\bibliographystyle{ieeetr}
\bibliography{main}
}
\medskip

\end{document}

%% file: sections/intro.tex
Adversarial examples have brought uncertainties and threatened the robustness and safety of learning-based decision-making systems, 
especially for autonomous driving \cite{NVIDIA}\cite{bojarski2016end}\cite{tesla} and robotic systems\cite{kumra2017robotic}\cite{mahler2017dex}\cite{lenz2015deep}, in which  perception is served as important inputs to the entire system. 
As an intriguing property first found in \cite{szegedy2013intriguing}, carefully computed perturbations on inputs could lead to misclassification with high confidence, even though the perturbations are imperceptible to humans. 
Over the past few years, researchers have developed a number of adversarial attacks \cite{goodfellow2014explaining}\cite{carlini2017towards}\cite{papernot2016limitations} on classification systems, which show the underlying instability for decision making with deep networks.
In later studies, noisy patterns are observed in the feature maps of adversarial examples \cite{xie2018feature}.
Motivated by the noisy pattern, we explore the possibility of defense with various smoothing techniques at test time. 
Instead of proving the superiority of a particular method, the focus of this paper is to reveal the adaptive characteristic of adversarial robustness. Meanwhile, the test-time smoothing methods discussed in this paper can still complement other defense schemes such as adversarial training \cite{goodfellow2014explaining}.

The contributions of this paper are: 
\begin{itemize}
    \item We implement a test-time defense pipeline which can smooth both the original inputs and intermediate outputs from any specified layer(s). This pipeline can be generalized to study the influence of intermediate layers in various tasks. 
    \item We present the non-monotonic relation between adversarial attacks and smoothing defenses: For a fixed attack, the successful defense rate first increases then decreases as smoothing becomes stronger. For a fixed defense, the classification accuracy on adversarial examples first drops then rebounds as the number of attack iterations increases. 
    \item We conduct the first investigation on the performance of defense on each test sample. The variance among samples is so large that it becomes easy to select large  (i.e., at scale $10^4$) subsets from ImageNet \cite{ILSVRC15} validation set, allowing a designated method to outperform others, or even to inflate $100\%$ accuracy.
    \item We demonstrate that different adversarial examples require different degrees of defense at test time and illustrate the potential benefits of iterative defenses. 
\end{itemize}
The rest of the paper is organized as follows. Section \ref{sec:literature} reviews related studies. Section \ref{sec:methodology} explains the methodology of test-time defense with various smoothing techniques. Sections \ref{sec:experiment} and \ref{sec:discussion} follow with experiments and discussion, respectively. 

All relevant materials that accompany this paper, including both the codes and pre-computed data, are publicly available at GitHub\footnote{\url{https://github.com/mkt1412/testtime-smoothing-defense}} and 
Dropbox\footnote{\url{https://www.dropbox.com/sh/ayiyiboba5nyv2k/AAAYZffyD0CeY_1aOmrkrg8Ba?dl=0}}.

%% file: sections/literature.tex
Since the discovery of adversarial examples, researchers have been working on methods of attacking or securing neural networks. We can roughly categorize the attacks with multiple criteria such as white-box or black-box, targeted or untargeted, and one-shot or iterative \cite{gomes18adversarial}.
On the defenders' side, two major strategies \cite{mustafa2019image} have been adopted in current practice: (1) reinforcing the classifier during training time so that it accounts for adversarial examples and (2) detecting and converting adversarial examples back to normal at test time. A comprehensive survey on adversarial examples is available in \cite{huang2018safety}. 
It is worth stating that fooling machine-learning algorithms with adversarial examples is much easier than designing models that cannot be fooled \cite{goodfellow2017attacking}. One can even fool a neural network with very few pixels \cite{su2019one}. For efficiency purpose, most existing attacks utilize gradient information of the network and ``optimally'' perturb the image to maximize the loss function. The noisy pattern might be a side-effect of such a gradient-based operation.

In this paper, we limit our scope to white-box untargeted attacks, which is the most common type in literature. Compared with the feature denoising in \cite{xie2018feature}, all our smoothing defenses are performed at test time. We assume that the neural-network classifier has already been shipped and deployed, or it might not be feasible to retrain with adversarial examples. In addition, we believe test-time defenses studied in this paper can also serve as a useful complimentary post-processing procedure, even when adversarial training is affordable. Contrary to existing work which often compares methods at dataset level with static configurations (e.g., a few sets of fixed parameters), we thoroughly investigate the behavior of test-time defenses at multiple levels and with varying strength. Smoothing methods are selected for illustration because their strength can be naturally measured using the number of iterations or radius of kernels.    

%% file: sections/methodology.tex
In this section, we elaborate the general test-time defense scheme and all relevant smoothing techniques that we experiment with. 
\subsection{Test-time defense scheme}
Let $f:X\rightarrow Y$ denote a pretrained classifier that maps an image $x \in X$ to its label $y \in Y$. An adversarial attack $g:X\rightarrow X$ then maps a legitimate image $x$ to an adversarial example $\hat{x}$ under certain constraints (e.g., on $L_p$ distance) such that $f(x) \neq f(\hat{x}) = f(g(x))$. To defend adversarial attacks at test time, an ideal solution would be applying the inverse mapping $g^{-1}: \hat{x}\mapsto x$. In reality, however, we have to find a defense $h$, which is an alternative approximation of $g^{-1}$ with the hope that $f(x) = f(h(\hat{x}))$ can be satisfied. In addition, a defense $h$ is more desirable for deployment if it brings less distortion to legitimate images $x$, keeping $f(x) = f(h(x))$. To achieve that, we may also introduce a detector (as a part of $h$) that distinguishes adversarial examples from legitimate ones at the first stage of the defense. Once an input is considered legitimate, no further defense is required. 

In this work, we apply smoothing techniques as the alternative approximation ($h$) of the inverse mapping ($g^{-1}$) of the attack. Theoretically, the smoothing defense only works when outputs from an adversarial attack ($g$) are ``noisy,'' which implies $g\approx h^{-1}$ from the perspective of $f$.

\subsection{Smoothing techniques}
The smoothing techniques involved can be categorized into three groups: common, edge-preserving, and advanced. The common group includes mean, median and Gaussian filter, which are most commonly used in image processing. Edge-preserving smoothing algorithms include anisotropic diffusion and bilateral filter. More advanced smoothing techniques include non-local means and modified curvature motion. We will explain concisely the algorithms in the following paragraphs. 

\textbf{Mean, median, and Gaussian filters}: These filters are widely applied in image processing. Despite the simple forms, they do not necessarily perform the worst in defending adversarial examples. 

\textbf{Anisotropic diffusion} \cite{perona1990scale}: The Perona-Malik anisotropic diffusion aims at reducing image noise without removing important edges by assigning lower diffusion coefficients for edge pixels (which have larger gradient norm). During iterations, the image is updated according to the formula below.  
$$I_t = \mathrm{div} \left( c(x,y,t) \nabla I \right)= \nabla c \cdot \nabla I + c(x,y,t) \Delta I$$
in which $\mathrm{div}$ denotes the divergence operator, $\Delta$ denotes the  Laplacian, and $\nabla$ denotes the gradient. 
The diffusion coefficient is defined either as
$c\left(\|\nabla I\|\right) = e^{-\left(\|\nabla I\| / K\right)^2}$ or 
$c\left(\| \nabla I\| \right) = \frac{1}{1 + \left(\|\nabla I\|/ K\right)^2}$.

\textbf{Bilateral filter} \cite{tomasi1998bilateral}: A bilateral filter is a non-linear edge-preserving filter that computes the filtered value of a pixel $p$ using weighted average of its neighborhood $S$. The filter can be expressed with the following formula \cite{paris2007gentle}. 
$$
\text{BF}[I](p) = \frac{1}{W(p)} \sum_{q\in S} G_{\sigma_s}(\|p - q\|) \, G_{\sigma_r}(|I(p) - I(q)|) \, I(q)
$$
in which $W(p) = \sum_{q\in S} G_{\sigma_s}(\|p - q\|) \, G_{\sigma_r}(|I(p) - I(q)|)$ is the normalization term. $G_{\sigma_s}$ and $G_{\sigma_r}$ are weight functions (e.g., Gaussian) for space and range, respectively. Edges are preserved because pixels that fall on different sides of the edge will have lower weights for range.

\textbf{Non-local means} \cite{buades2005non}: The non-local mean algorithm takes a more general form in which the output value of a pixel $i$ is computed as a average of all pixels in the image $I$, weighted by a similarity $w(i, j)$ which is measured as a decreasing function of the weighted Euclidean distance to that pixel. For a discrete image $v = \{v(i)\,|\,i \in I\}$, the filtered value for a pixel $i$ is computed as follows. $$
\text{NL}[v](i) = \sum_{j\in I} w(i,j) v(i)
$$
in which the weights 
$
w(i,j) = \frac{1}{Z(i)} e^{-\frac{\|v(\mathcal{N}_i) - v(\mathcal{N}_j)\|^2_{2, a}}{h^2}}
$. In the formula, $\mathcal{N}_k$ denotes a square neighborhood of fixed size centered at a pixel $k$, and $a$ is the standard deviation of the Gaussian kernel. The normalizing constant is computed as $Z(i) = \sum_j e^{-\frac{\|v(\mathcal{N}_i) - v(\mathcal{N}_j)\|^2_{2, a}}{h^2}}$.

\textbf{Modified curvature motion} \cite{yezzi1998modified}: As most smoothing techniques are originally designed for gray-scale images, generalizing them to multi-channel color images and feature maps might be less natural, and sometimes there may exist multiple ways for the generalization.  
Instead of splitting a color image into separate channels, we can treat it as a surface $(x, y, R(x,y), G(x,y), B(x,y)) \subset \mathds{R}^5$. Following the geometric property that smoother surfaces have smaller areas (or volumes), we can then iteratively smooth it with a general curvature motion method: 
$$
I_t = \frac{k^{-2}\nabla^2 I + (I_y^2 + I_z^2)I_{xx} + (I_x^2 + I_z^2)I_{yy} + (I_x^2 + I_y^2)I_{zz} - 2(I_x I_y I_{xy} + I_x I_z I_{xz} + I_y I_z I_{yz})}{(k^{-2} + \|\nabla I\|^2)^2}
$$ 
where $k$ is a scaling factor. As $k$ becomes larger, the algorithm transits from isotropic to a more edge-preserving diffusion.  
Such a formulation can be easily and naturally extended to feature maps with more channels along the $z$ axis. 

%% file: sections/experiment.tex
In this section, we present our experimental results of test-time smoothing against adversarial examples.
In order to prepare the test set, we first select all images (39,156 in total) that are correctly classified by the pretrained ResNet-152 \cite{he2016deep} . Then we generate and store adversarial examples using attacks that are provided by Foolbox \cite{rauber2017foolbox} and ART \cite{art2018}. The white-box untargeted attacks include Projected Gradient Descent (PGD) \cite{madry2017towards}, Deep Fool\cite{moosavi2016deepfool}, Saliency Map\cite{papernot2016limitations}, Newton Fool \cite{jang2017objective},
and salt-and-pepper noise. 
For strong attacks (e.g., PGD) that cannot be mitigated by quantization, we store the adversarial examples in \emph{jpeg} format; for other attacks that require adversarial examples in floating point accuracy, we store their results in \emph{pkl} files. To save computation time for future work, all generated data will be available for download. 
\subsection{Performance of various smoothing techniques on defending fixed adversarial attacks} \label{subsec:fixed-attack_various-defenses}
We start with a set of controlled experiments that show the performance of various smoothing techniques on defending a fixed attack. 
Two sets of PGD attacks, with maximum perturbation $\epsilon=0.01$ (imperceptible to humans) and $\epsilon = 0.05$ (similar scale as in \cite{xie2018feature}), are chosen as baseline because (1) PGD attack is one of the strongest attacks and (2) adversarial examples from PGD attack cannot be defended by simple quantization. 
Following the default settings, 20 iterations of PGD are performed. 
After obtaining the perturbed images, we tweak the parameters in each smoothing method to pursue the optimal ones that lead to the highest classification accuracy over all perturbed images. If a method contains multiple parameters, we tweak them one after another and naively apply the optimal parameter values obtained from previous exploration. 

\begin{figure}
\centering  
\subfigure[PGD ($\epsilon=0.01$, 20 iterations)]{
\includegraphics[width=0.45\textwidth]{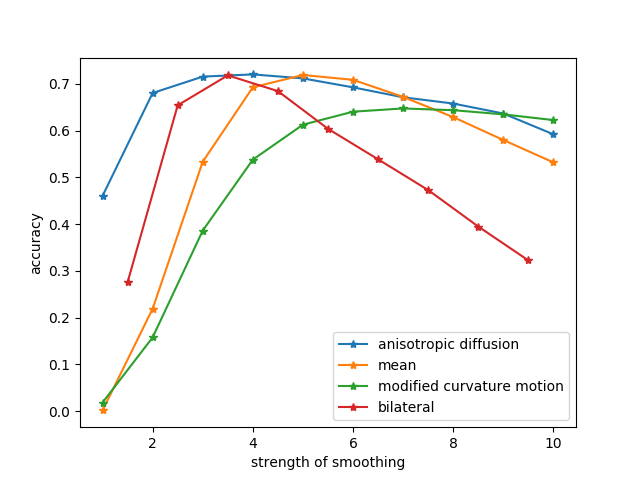}}
\subfigure[PGD ($\epsilon=0.05$, 20 iterations)]{
\includegraphics[width=0.45\textwidth]{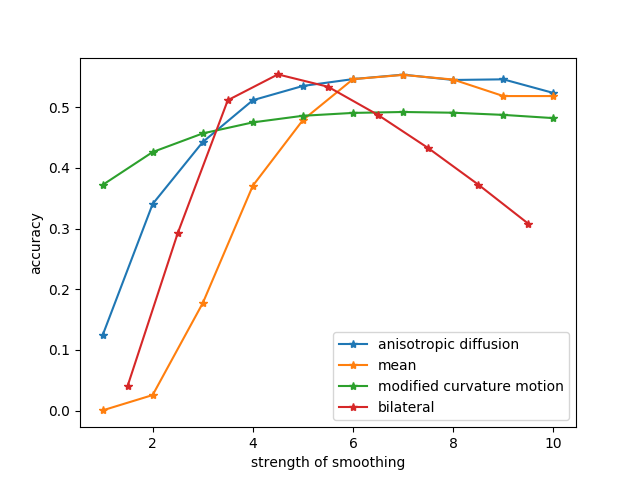}}
\caption{\label{fig:fixed-attack_various-defenses} Change of classification accuracy on ImageNet validation set (vertical) along with the strength of smoothing defense (horizontal). The strength of smoothing method is measured by the most sensitive parameter: number of iterations for anisotropic diffusion and modified curvature motion, size of the kernel for mean filter, and radius of the neighborhood for bilateral filter.}
\end{figure}

Figure \ref{fig:fixed-attack_various-defenses} shows the classification accuracy on ImageNet validation set as the strength of smoothing defenses varies. The strength is measured by the most sensitive parameter, that is, \emph{number of iterations} for iterative methods such as anisotropic diffusion and modified curvature motion, \emph{size of the kernel} for mean filters, and \emph{diameter of the neighborhood} for bilateral filters. 
We only present results from four selected methods because the rest of them lead to much lower (i.e., 20-30\% less) classification accuracy. Henceforth, we will focus on these four methods in subsequent sections. 

\begin{figure}
    \centering
    \subfigure[fixed attack $g$, varying defense $h$ \label{subfig:detour_fixed-attack_varying-defense}]{
\includegraphics[width=0.35\textwidth]{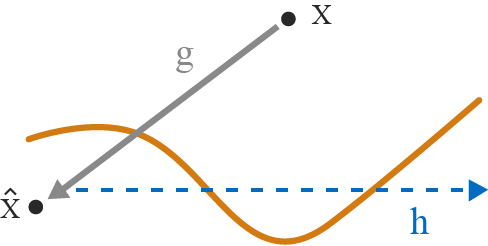}}
    \quad
    \subfigure[varying attack $g$, fixed defense $h$ \label{subfig:detour_varying-attack_fixed-defense}]{
\includegraphics[width=0.35\textwidth]{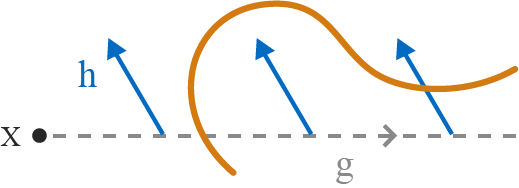}}
    \caption{Simplified illustration on the ``detour'' effect between adversarial attack $g$ and test-time defense $h$}
    \label{fig:detour}
\end{figure}
The curves in Figure \ref{fig:fixed-attack_various-defenses} share a similar concave shape, which might suggest a geometric relation between adversarial attack $g$ and test-time defense $h$.  
As illustrated by Figure \ref{subfig:detour_fixed-attack_varying-defense}, the test-time defense should not travel too far along the ``detour.'' In the following subsection, we further study the non-monotonic effect from the attackers' perspective. 
\subsection{Performance of a fixed defense under attacks with varying number of iterations} \label{subsec:varying-attack_fixed-defense}
\begin{figure}
\centering  
\subfigure[PGD ($\epsilon=0.01$)]{
\includegraphics[width=0.45\textwidth]{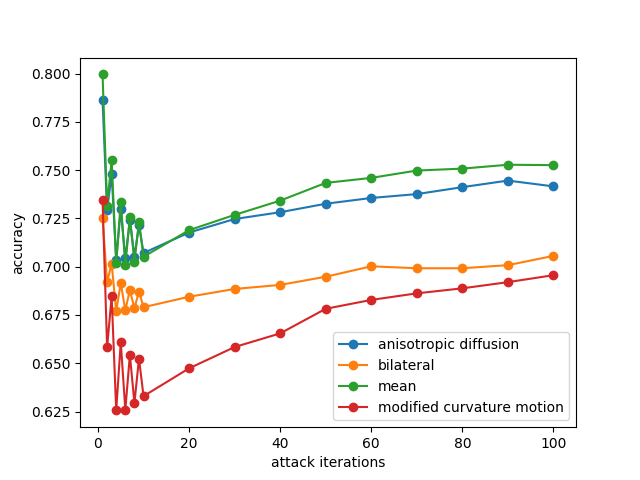}}
\subfigure[PGD ($\epsilon=0.05$)]{
\includegraphics[width=0.45\textwidth]{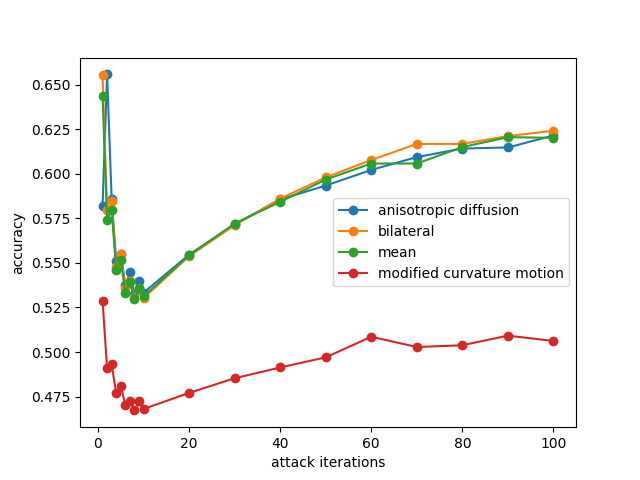}}
\caption{\label{fig:varying-attack_fixed-defenses} Change of classification accuracy on ImageNet validation set (vertical) along with the number of iterations in PGD attack (horizontal). The bump at iteration $=50$ corresponds to a switch from the ImageNet validation set to a subset of 5,000 images to reduce computation time.}
\end{figure}

We continue our controlled experiments by setting the parameters of each smoothing defense to the optimal values obtained in section \ref{subsec:fixed-attack_various-defenses} and varying the strength (i.e., number of iterations) of PGD attacks. 
Figure \ref{fig:varying-attack_fixed-defenses} presents the classification accuracy on ImageNet validation set as the number of attack iteration increases from 1 to 100. 
Surprisingly, the accuracy first drops but then rebounds as the number of attack iterations keeps increasing. Such performance might seem contradictory to previous work as we used to believe that more iterations leads to stronger attacks, especially for defenses that involve adversarial training. For test-time defenses, however, the convex curves may reflect the actual non-monotonic relation between attacks $g$ and defenses $h$, as illustrated in Figure \ref{subfig:detour_varying-attack_fixed-defense}. In contrast, when no defense is performed, the classification accuracy keeps dropping and stabilizes at a low level.
\subsection{Variance of performance among categories} \label{subsec:variance-among-categories}
\begin{figure}
\centering  
\subfigure[PGD, $\epsilon=0.01$ \label{subfig:categorical-accuracy_pgd-0.01}]{
\includegraphics[width=0.4\textwidth, trim={0 0 0 1.2cm}, clip]{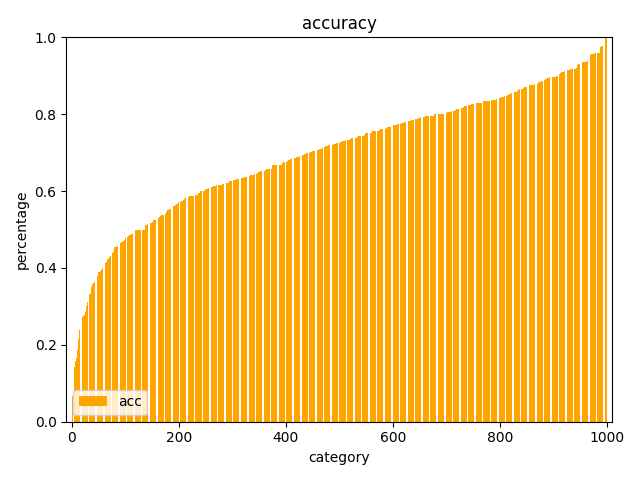}}
\subfigure[PGD, $\epsilon=0.05$]{
\includegraphics[width=0.4\textwidth, trim={0 0 0 1.2cm}, clip]{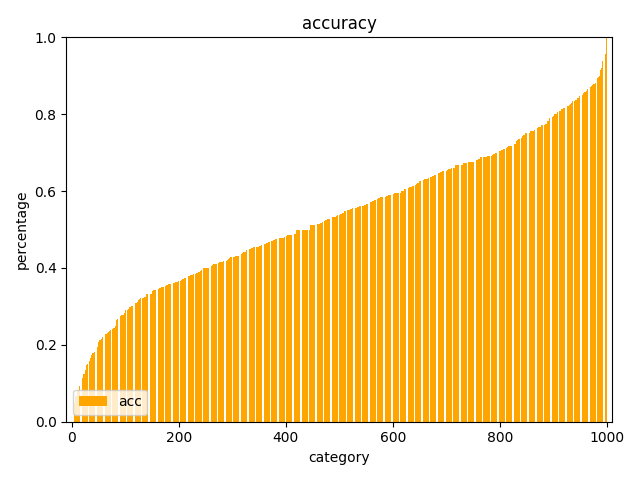}}
\caption{\label{fig:variance-among-categories} Distribution of categorical accuracy on adversarial examples in increasing order. (a): Results on adversarial examples that are generated from PGD ($\epsilon=0.01$), with anisotropic diffusion as defense. (b): same as (a) but the adversarial examples are generated from PGD ($\epsilon=0.05$).} 
\end{figure}

During the experiments, we noticed that the variance of classification accuracy for each category was quite large. For illustration purposes, we take PGD attack and anisotropic diffusion as an example.  
Figure \ref{subfig:categorical-accuracy_pgd-0.01} shows the sorted accuracies from ImageNet categories. The lowest categorical accuracy stays below 20\% whereas the highest accuracy reaches almost 100\%. Similar distribution of categorical accuracy is observed from other attack-defense pairs. 
Such an observation leads us to a question: is it possible to select a relatively large subset of test samples on which a designated method works the best? 
The task turns out to be easy.  
For each smoothing technique, we sort the test samples that are correctly classified according to their prediction confidence. Then, we can select a relatively large (with more than 20,000 samples) "optimal" subset with high prediction confidence. 
\begin{table}
  \caption{\label{tab:optimal-subsets}Classification accuracy on ``optimal'' subsets consisting of adversarial examples generated from PGD attack ($\epsilon=0.01$). Accuracies can be inflated to 100\% on a dataset with $>10^4$ samples.}
  \centering
  \begin{tabular}{|l|ccc|}
\hline
\diagbox{"Optimal" subset }{Defense} & Anisotropic diffusion & Bilateral & Mean \\
\hline
Anisotropic diffusion & 100.0\% & 88.51\% &93.79\% \\
Bilateral & 92.79\%  & 100\% & 93.81\% \\
Mean  &93.62\% &89.31\% &100\% \\
\hline
\end{tabular}
\end{table}
The performance on these optimal subsets are shown in Table \ref{tab:optimal-subsets}. For example, the subset at the first row is selected based on anisotropic diffusion. 
Therefore, anisotropic diffusion achieves 100\% accuracy while bilateral filters only achieve less than 90\%. 
The optimal subset for each smoothing technique in this work will be released. 
\subsection{Variance of required defense for test samples} \label{subsec:variance-among-samples}
\begin{figure}
\centering  
\subfigure[PGD ($\epsilon=0.01$) minimum iteration]{
\includegraphics[width=0.45\textwidth]{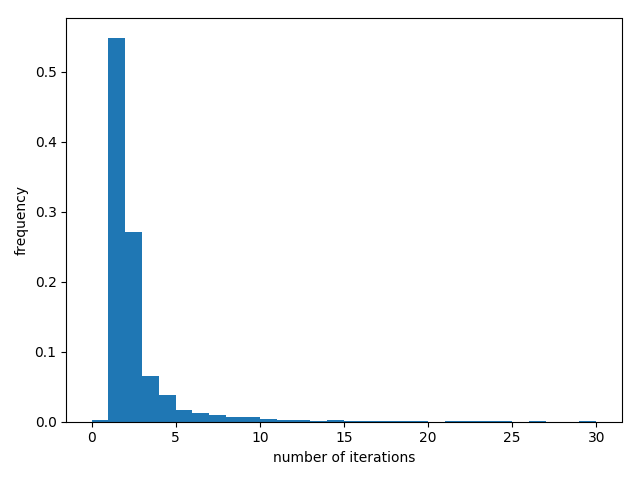}}
\subfigure[PGD ($\epsilon=0.05$) minimum iteration]{
\includegraphics[width=0.45\textwidth]{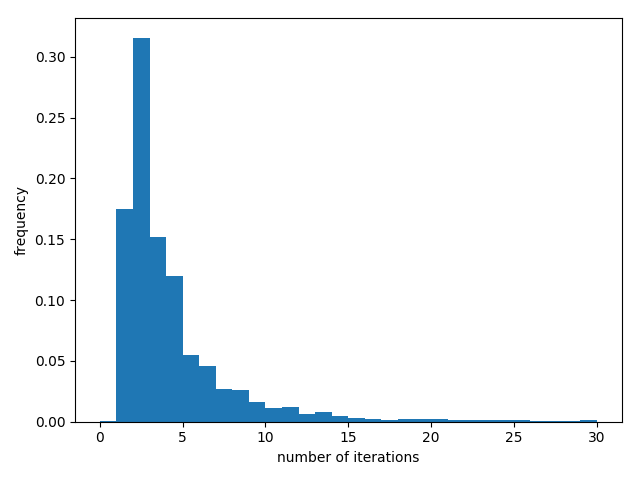}}
\caption{Minimum number of iterations required for defending adversarial examples}
\label{fig:min-iteration}
\end{figure}

Both non-monotonic relations in section \ref{subsec:fixed-attack_various-defenses} and large variance in section \ref{subsec:variance-among-categories} suggest the idea of an adaptive version of the test-time smoothing defense, which is favorable for iterative methods. Specifically, an optimal iteration number or termination criterion is required for each adversarial example.  
In order to demonstrate the potential advantage of the adaptive method, we compute the minimum number of iterations required for defending an adversarial example. Figure \ref{fig:min-iteration} shows the histograms of minimum iterations required with anisotropic diffusion under two sets of PGD attacks. In addition, the upper bound of the minimum iteration number is set to 30. In other words, if an adversarial sample remains misclassified throughout 30 smoothing iterations, we consider it as undefendable. We then compute an upper-bound accuracy for the defense by taking account results from all iterations. Compared with the result from a fixed iteration number over the  whole dataset, our simulation of the “adaptive method” enhance the accuracy from 72.2\% to 83.6\% on adversarial examples generated by PGD ($\epsilon=0.01$) and from 55.5\% to 70.1\% on adversarial examples generated by PGD ($\epsilon=0.05$).
Designing and implementing the adaptive algorithm is left for future work. 

%% file: sections/discussion.tex
In this work, we present the non-monotonic relation between adversarial attacks and test-time defenses. 
The huge variance of classification accuracy on adversarial samples may cast doubts on previous work with unpublished or relatively small datasets, as selecting a large-scale dataset to inflate accuracy for a particular defense is feasible. For this reason, we did not argue the superiority of a single defense. 
As already stated in previous sections, the effectiveness of test-time smoothing defenses rely on a strong assumption that adversarial images are more noisy than clean samples. Such an assumption, however, may not be valid for more advanced attacks in the future. Based on results from our experiment, test-time smoothing is helpful in defending exisiting white-box untargeted attacks. 
There are a few failed attempts that are worth reporting.  Edge-preserving smoothing techniques do not necessarily outperform common filters because they may also preserve structural details in strong noise. In addition, multi-channel smoothing techniques are not superior to channel-by-channel smoothing filters. Different from the feature denoising at adversarial training, feature smoothing at test time becomes less useful once the smoothing is sufficient at the image domain. 

We hope this paper will stimulate more detailed analysis on adversarial examples at a finer scale. Future work includes investigating and comparing characteristics of defendable and undefendable adversarial examples, as well as constructing robust and adaptive termination criteria for each test sample. In addition, a more comprehensive metric is needed for the evaluation of defenses.

%% file: main.bbl
\begin{thebibliography}{10}

\bibitem{NVIDIA}
``Nvidia self driving vehicles development platform..''
  \url{http://www.nvidia.com/object/drive-px.html.}
\newblock Accessed: 2019-09-17.

\bibitem{bojarski2016end}
M.~Bojarski, D.~Del~Testa, D.~Dworakowski, B.~Firner, B.~Flepp, P.~Goyal, L.~D.
  Jackel, M.~Monfort, U.~Muller, J.~Zhang, {\em et~al.}, ``End to end learning
  for self-driving cars,'' {\em arXiv preprint arXiv:1604.07316}, 2016.

\bibitem{tesla}
``Tesla. autopilot | tesla..'' \url{https://www.tesla.com/autopilot.}
\newblock Accessed: 2019-09-17.

\bibitem{kumra2017robotic}
S.~Kumra and C.~Kanan, ``Robotic grasp detection using deep convolutional
  neural networks,'' in {\em 2017 IEEE/RSJ International Conference on
  Intelligent Robots and Systems (IROS)}, pp.~769--776, IEEE, 2017.

\bibitem{mahler2017dex}
J.~Mahler, J.~Liang, S.~Niyaz, M.~Laskey, R.~Doan, X.~Liu, J.~A. Ojea, and
  K.~Goldberg, ``Dex-net 2.0: Deep learning to plan robust grasps with
  synthetic point clouds and analytic grasp metrics,'' {\em arXiv preprint
  arXiv:1703.09312}, 2017.

\bibitem{lenz2015deep}
I.~Lenz, H.~Lee, and A.~Saxena, ``Deep learning for detecting robotic grasps,''
  {\em The International Journal of Robotics Research}, vol.~34, no.~4-5,
  pp.~705--724, 2015.

\bibitem{szegedy2013intriguing}
C.~Szegedy, W.~Zaremba, I.~Sutskever, J.~Bruna, D.~Erhan, I.~Goodfellow, and
  R.~Fergus, ``Intriguing properties of neural networks,'' {\em arXiv preprint
  arXiv:1312.6199}, 2013.

\bibitem{goodfellow2014explaining}
I.~J. Goodfellow, J.~Shlens, and C.~Szegedy, ``Explaining and harnessing
  adversarial examples,'' {\em arXiv preprint arXiv:1412.6572}, 2014.

\bibitem{carlini2017towards}
N.~Carlini and D.~Wagner, ``Towards evaluating the robustness of neural
  networks,'' in {\em 2017 IEEE Symposium on Security and Privacy (SP)},
  pp.~39--57, IEEE, 2017.

\bibitem{papernot2016limitations}
N.~Papernot, P.~McDaniel, S.~Jha, M.~Fredrikson, Z.~B. Celik, and A.~Swami,
  ``The limitations of deep learning in adversarial settings,'' in {\em 2016
  IEEE European Symposium on Security and Privacy (EuroS\&P)}, pp.~372--387,
  IEEE, 2016.

\bibitem{xie2018feature}
C.~Xie, Y.~Wu, L.~van~der Maaten, A.~Yuille, and K.~He, ``Feature denoising for
  improving adversarial robustness,'' {\em arXiv preprint arXiv:1812.03411},
  2018.

\bibitem{ILSVRC15}
O.~Russakovsky, J.~Deng, H.~Su, J.~Krause, S.~Satheesh, S.~Ma, Z.~Huang,
  A.~Karpathy, A.~Khosla, M.~Bernstein, A.~C. Berg, and L.~Fei-Fei, ``{ImageNet
  Large Scale Visual Recognition Challenge},'' {\em International Journal of
  Computer Vision (IJCV)}, vol.~115, no.~3, pp.~211--252, 2015.

\bibitem{gomes18adversarial}
J.~Gomes, ``Adversarial attacks and defences for convolutional neural
  networks,'' 2018.

\bibitem{mustafa2019image}
A.~Mustafa, S.~H. Khan, M.~Hayat, J.~Shen, and L.~Shao, ``Image
  super-resolution as a defense against adversarial attacks,'' {\em arXiv
  preprint arXiv:1901.01677}, 2019.

\bibitem{huang2018safety}
X.~Huang, D.~Kroening, M.~Kwiatkowska, W.~Ruan, Y.~Sun, E.~Thamo, M.~Wu, and
  X.~Yi, ``Safety and trustworthiness of deep neural networks: A survey,'' {\em
  arXiv preprint arXiv:1812.08342}, 2018.

\bibitem{goodfellow2017attacking}
A.~Goodfellow and B.~Papernot, ``Is attacking machine learning easier than
  defending it?,'' {\em cleverhans-blog}, vol.~15, 2017.

\bibitem{su2019one}
J.~Su, D.~V. Vargas, and K.~Sakurai, ``One pixel attack for fooling deep neural
  networks,'' {\em IEEE Transactions on Evolutionary Computation}, 2019.

\bibitem{perona1990scale}
P.~Perona and J.~Malik, ``Scale-space and edge detection using anisotropic
  diffusion,'' {\em IEEE Transactions on pattern analysis and machine
  intelligence}, vol.~12, no.~7, pp.~629--639, 1990.

\bibitem{tomasi1998bilateral}
C.~Tomasi and R.~Manduchi, ``Bilateral filtering for gray and color images.,''
  in {\em Iccv}, vol.~98, p.~2, 1998.

\bibitem{paris2007gentle}
S.~Paris, P.~Kornprobst, J.~Tumblin, and F.~Durand, ``A gentle introduction to
  bilateral filtering and its applications,'' in {\em ACM SIGGRAPH 2007
  courses}, p.~1, ACM, 2007.

\bibitem{buades2005non}
A.~Buades, B.~Coll, and J.-M. Morel, ``A non-local algorithm for image
  denoising,'' in {\em 2005 IEEE Computer Society Conference on Computer Vision
  and Pattern Recognition (CVPR'05)}, vol.~2, pp.~60--65, IEEE, 2005.

\bibitem{yezzi1998modified}
A.~Yezzi, ``Modified curvature motion for image smoothing and enhancement.,''
  {\em IEEE transactions on image processing: a publication of the IEEE Signal
  Processing Society}, vol.~7, no.~3, pp.~345--352, 1998.

\bibitem{he2016deep}
K.~He, X.~Zhang, S.~Ren, and J.~Sun, ``Deep residual learning for image
  recognition,'' in {\em Proceedings of the IEEE conference on computer vision
  and pattern recognition}, pp.~770--778, 2016.

\bibitem{rauber2017foolbox}
J.~Rauber, W.~Brendel, and M.~Bethge, ``Foolbox: A python toolbox to benchmark
  the robustness of machine learning models,'' {\em arXiv preprint
  arXiv:1707.04131}, 2017.

\bibitem{art2018}
M.-I. Nicolae, M.~Sinn, M.~N. Tran, A.~Rawat, M.~Wistuba, V.~Zantedeschi,
  N.~Baracaldo, B.~Chen, H.~Ludwig, I.~Molloy, and B.~Edwards, ``Adversarial
  robustness toolbox v0.8.0,'' {\em CoRR}, vol.~1807.01069, 2018.

\bibitem{madry2017towards}
A.~Madry, A.~Makelov, L.~Schmidt, D.~Tsipras, and A.~Vladu, ``Towards deep
  learning models resistant to adversarial attacks,'' {\em arXiv preprint
  arXiv:1706.06083}, 2017.

\bibitem{moosavi2016deepfool}
S.-M. Moosavi-Dezfooli, A.~Fawzi, and P.~Frossard, ``Deepfool: a simple and
  accurate method to fool deep neural networks,'' in {\em Proceedings of the
  IEEE conference on computer vision and pattern recognition}, pp.~2574--2582,
  2016.

\bibitem{jang2017objective}
U.~Jang, X.~Wu, and S.~Jha, ``Objective metrics and gradient descent algorithms
  for adversarial examples in machine learning,'' in {\em Proceedings of the
  33rd Annual Computer Security Applications Conference}, pp.~262--277, ACM,
  2017.

\end{thebibliography}
